\def\year{2015}
\documentclass[letterpaper]{article}
\usepackage{aaai}
\usepackage{times}
\usepackage{helvet}
\usepackage{courier}
\usepackage{graphicx}
\usepackage{subfigure}
\usepackage[super]{nth}
\usepackage{amssymb}
\usepackage{url}

\usepackage{epstopdf}

\begin{document}
%
\title{On Vectorization of Deep Convolutional Neural Networks for Vision Tasks}
\author{Jimmy SJ. Ren ~~~~~ Li Xu\\
Lenovo Research \& Technology \\
\url{http://vcnn.deeplearning.cc}\\
\url{jimmy.sj.ren@gmail.com}~~~~~\url{xulihk@lenovo.com}}

\maketitle
\begin{abstract}
\begin{quote}
We recently have witnessed many ground-breaking results in machine
learning and computer vision, generated by using deep convolutional
neural networks (CNN). While the success mainly stems from the large
volume of training data and the deep network architectures, the
vector processing hardware (e.g. GPU) undisputedly plays a vital
role in modern CNN implementations to support massive computation.
Though much attention was paid in the extent literature to
understand the algorithmic side of deep CNN, little research was dedicated to the
vectorization for scaling up CNNs. In this paper, we studied the
vectorization process of key building blocks in deep CNNs, in
order to better understand and facilitate parallel implementation. Key steps in training
and testing deep CNNs are abstracted as matrix and vector operators,
upon which parallelism can be easily achieved. We developed and
compared six implementations with various degrees of vectorization
with which we illustrated the impact of vectorization on the speed
of model training and testing. Besides, a unified CNN framework for
both high-level and low-level vision tasks is provided, along with a
vectorized Matlab implementation with state-of-the-art speed performance.
\end{quote}
\end{abstract}

\section{Introduction}
\noindent Deep convolutional neural network (CNN) has become a keen
tool in addressing large scale artificial intelligence tasks. Though
the study of CNN can be traced back to late 1980s
\cite{lecun89,lecun90}, the recent success of deep CNN is largely
attributed to the concurrent progresses of the two technical
streams. On the one hand, the new deep CNN architecture with
elements such as Dropout \cite{hinton12,KrizhevskySH12}, DropConnect
\cite{WanZZLF13}, Rectified Linear Units-ReLU \cite{NairH10} as well
as new optimization strategies \cite{DeanCMCDLMRSTYN12} have
empowered deep CNN with greater learning capacity. On the other
hand, the rapid advances and democratization of high performance
general purpose vector processing hardware, typified by graphics
processing unit (GPU), unleashes the potential power of deep CNN by
scaling up the network significantly.


Various infrastructures were used in scaling up deep CNNs, including
GPU \cite{CoatesHWWCN13}, distributed CPU based framework
\cite{DeanCMCDLMRSTYN12}, FPGA \cite{farabet09}, etc. Though the
implementation details among those approaches differ, the core
insight underlying the idea of scaling up deep CNN is
parallelization \cite{bengio07} in which vectorization technique is
the fundamental element. While the consecutive distinguished
performance of GPU trained CNNs in the ImageNet visual recognition
challenge \cite{KrizhevskySH12,RussakovskyDHBF13} as well as the
reported results in many studies in the literature justify its
effectiveness \cite{jia14,sermanet13}, the published literature did
not provide sufficient insights on how the vectorization was carried
out in detail. We also found there is no previous study to answer
how different degrees of vectorization influence the performance of
deep CNN, which is, however, crucial in finding the bottlenecks and
helps to scale up the network architecture. We believe these
questions form a significant research gap and the answer to these
questions shall shed some light on the design, tuning and
implementation of vectorized CNNs.
%



In this paper, we reinterpret the key operators in deep CNNs in vectorized forms with which high parallelism can be easily achieved given basic parallelized matrix-vector operators. To show the impact of the vectorization on the speed of both model training and testing, we developed and compared six implementations of CNNs with various degrees of vectorization. We also provide a unified framework for both high-level and low-level vision applications including recognition, detection, denoise and image deconvolution. Our Matlab Vectorized CNN implementation (VCNN) will be made publicly available on the project webpage.

\subsection{Related Work}

\noindent Efforts on speeding up CNN by vectorization starts with
its inception. Specialized CNN chip \cite{jackel90} was built and
successfully applied to handwriting recognition in the early 90s.
Simard et al. \shortcite{simard03} simplified CNN by fusing
convolution and pooling operations. This speeded up the network and
performed well in document analysis. Chellapilla et al.
\shortcite{chellapilla06} adopted the same architecture but unrolled
the convolution operation into a matrix-matrix product. It has now
been proven that this vectorization approach works particularly well
with modern GPUs. However, limited by the available computing power,
the scale of the CNN explored at that time was much smaller than
modern deep CNNs.

When deep architecture showed its ability to effectively learn
highly complex functions \cite{hinton06}, scaling up neural network
based models was soon becoming one of the major tasks in deep
learning \cite{bengio07}. Vectorization played an important role in
achieving this goal. Scaling up CNN by vectorized GPU
implementations such as Caffe \cite{jia14}, Overfeat
\cite{sermanet13}, CudaConvnet \cite{KrizhevskySH12} and Theano
\cite{bergstra10} generates state-of-the-art results on many vision
tasks. Albeit the good performance, few of the previous papers
elaborated on their vectorization strategies. As a consequence, how
vectorization affects design choices in both model training and
testing is unclear.

Efforts were also put in the acceleration of a part of the deep CNN
from algorithmic aspects, exemplified by the separable kernels for
convolution \cite{denton14} and the FFT speedup
\cite{mathieu2013fast}. Instead of finding a faster alternative for
one specific layer, we focus more on the general vectorization
techniques used in {\it all} building blocks in deep CNNs, which is
instrumental not only in accelerating existing networks, but also in
providing guidance for implementing and designing new CNNs across different platforms, for
various vision tasks.
%

\section{Vectorization of Deep CNN}
\noindent Vectorization refers to the process that transforms the
original data structure into a vector representation so that the
scalar operators can be converted into a vector implementation. In this section, we introduce vectorization strategies for different layers in Deep CNNs.

\begin{figure}[t]
  \centering
  \includegraphics[width=1.0\linewidth]{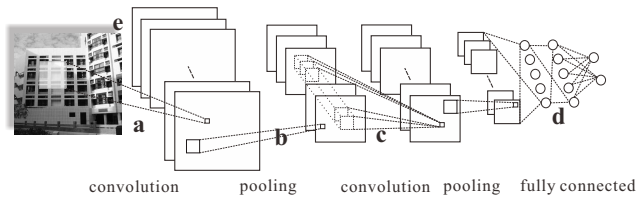}
  \caption{Convolutional Neural Network architecture for visual recognition.}
  \label{fig:fig1} 
\end{figure}

Figure \ref{fig:fig1} shows the
architecture of a typical deep CNN for vision tasks. It contains all of the essential parts of modern CNNs. Comprehensive introductions on
CNN's general architecture and the recent advances can be found in
\cite{lecun98} and \cite{KrizhevskySH12}.

We mark the places where vectorization plays
an important role. ``{\bf a}" is the convolution layer that
transforms the input image into feature representations, whereas
``{\bf b}" is the one to handle the pooling related operations.
``{\bf c}" represents the convolution related operations for feature
maps. We will see shortly that the vectorization strategies between
``{\bf a}" and ``{\bf c}" are slightly different. ``{\bf d}" involves operations in the fully connected network. Finally, ``{\bf e}" is the
vectorization operation required to simultaneously process multiple input samples
(e.g. mini-batch training). It is worth noting that we need to
consider both forward pass and back-propagation for all these
operations.

\subsection{Vectorizing Convolution}
\label{sec:conv_vect} We refer to the image and intermediate feature maps as $f$, one of the
convolution kernels as $w_i$, the convolution layer can be typically
expressed as
\begin{eqnarray}
f^{l+1}_i = \sigma(w^l_i*f^{l}+b_i^l),
\end{eqnarray}
where $i$ indexes the $i^{th}$ kernel. $l$ indexes the layer. $b^l$ is the bias weight.  $*$ is the convolution operator. For vision tasks, $f$ can be 2- or 3-dimension. The outputs from previous layer can be deemed as one single input $f^{l}$. $\sigma$ is the nonlinear function which could be ReLU, hyperbolic tangent, sigmoid, etc. Adding bias weight and applying nonlinear mapping are element-wise operations which can be deemed as already fully vectorized, i.e. the whole feature vector can be processed simultaneously. Contrarily, the convolution operators involve a bunch of multiplication with conflict memory access. Even the operators are parallized for each pixel, the parallelism \cite{ragan13} to be exploited is rather limited: compared to the number of computing units on GPU, the number of convolution in one layer is usually smaller. A fine-grained parallelism on element-wise multiplication is much preferred, leading to the vectorization process to unroll the convolution.

In what follows, all the original data $f$, $b$ and $w$ can be viewed as data vectors. Specifically, we seek vectorization operators $\varphi_c()$  to map kernel or feature map to its matrix form so that convolution can be conducted by matrix-vector multiplication. However, a straight forward kernel-matrix, image-vector
product representation of convolution is not applicable here, since the kernel matrix is a sparse block-Toeplitz-Toeplitz-block one, not suitable for
parallelization due to the existence of many zero elements. Thanks to the duality of kernel and feature map in convolution, we can construct a {\it dense} feature-map-matrix and a kernel-vector. Further, multiple kernels can be put together to form a matrix so as to generate multiple feature map outputs simultaneously,
\begin{eqnarray}
[f_i^{l+1}]_{i} = \sigma(\varphi_c(f^{l})[w_i^l]_i+[b_i^l]_i).
\end{eqnarray}
Operator $[~]_i$ is to assemble vectors with index $i$ to form a matrix.

\begin{figure}[t]
  \centering
  \includegraphics[width=1.0\linewidth]{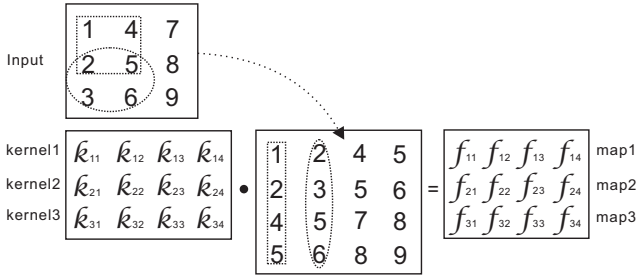}
  \caption{Strategy to vectorize convolution. Illustration of the way to covolve a 3x3 input image with three 2x2 kernels and generates three 2x2 feature maps.}
  \label{fig:fig2} 
\end{figure}

{\bf Backpropagation} The training procedure requires the backward propagation of gradients through $\varphi_c(f^{l})$. Note that $\varphi_c(f^{l})$ is in the unrolled matrix form, different form the outputs of previous layer $[f^{l}_i]_i$. An inverse operator $\varphi_c^{-1}()$ is thus required to transform the matrix-form gradients into the vector form for further propagation.  Since $\varphi_c()$ is a one-to-many mapping, $\varphi_c^{-1}()$ is a many-to-one operator. Fortunately, the gradient update is a linear process which can also be processed separately and combined afterwards.

{\bf Matlab Practice} Our matlab implementation to vectorize the input image is shown in Fig. \ref{fig:fig2}. Specifically, we first crop the image patches base on the kernel size and reorganize
them into columns, as indicated by the dotted bounds. Convolution kernels are
arranged by rows in another matrix. We can see that the product of
these two matrices will put all the convolved feature maps in the
resulting matrix, one feature map per row.  $\varphi_c()$ here can
be efficiently implemented by the \textit{im2col()} functions in
Matlab on both GPU \footnote{We used a custom version of
\textit{im2col()} for GPU.} and CPU. We note that the feature map vector here are transpose of $f_i$, simply because the function of \textit{im2col()}.

\begin{figure}[t]
  \centering
  \includegraphics[width=1.0\linewidth]{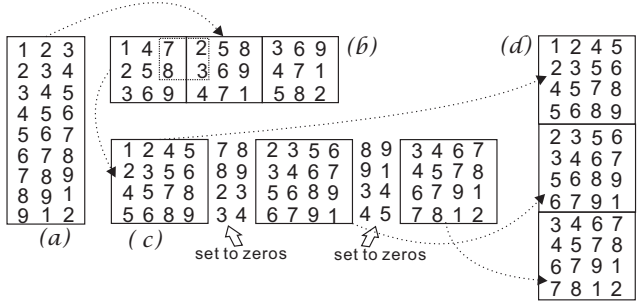}
  \caption{Strategy to vectorize convolution with a feature map. Illustration of the way to convolve a 3x3x3 feature map.}
  \label{fig:fig3} 
\end{figure}

\paragraph{\it Convolution with the feature map } An alternative practice is needed to
handle convolution of the feature map (e.g.
``{\bf c}" in Fig. \ref{fig:fig1}). This is because we need to first combine $[f^{l}_i]_i$ to a higher dimensional $f^{l}$ and then perform convolution.
One example is illustrated in Fig. \ref{fig:fig3}, where we need to combine 3 $3\times3$ feature maps to a $3\times3\times3$ one and apply $\varphi_c()$. In practice, we could first apply three times $\varphi_c()$ to $f^{l}_i$ and then combine the results. We found it less efficient since the actual number of feature map is much larger. To exploit more parallelism, we try to reorganize the data and apply only once the vectorization operator $\varphi_c()$.

%

In Fig. \ref{fig:fig3}, we first reshape column vectors $[f_i]_i$ (a) back to a 2D matrix
and put them side by side (b). This operation is cheap
since it just changes the representation of the data but does not
rearrange it. It allows us to vectorize all the feature
maps in a holistic way (c). Since only valid region is considered during convolution, we set all the redundant columns to zeros. The final $\varphi_c(f)$ is obtained by rearranging the intermediate result in (c). Since redundant columns are involved, we use Matlab function \textit{accumarray()} to handle many-to-one rearrangement.

One may note that the operator $\varphi^{-1}_c()$ is a many-to-one mapping as well. So it can be efficiently implemented by \textit{accumarray()}, for backpropagation.

\subsection{Vectorizing Pooling}

It is inefficient to carry out the pooling separately for each
feature map, so the goal here is to simultaneously process those
separate operations by vectorization. The pooling operator can be abstracted as
\begin{eqnarray}
f^{l+1} = \sigma(\varphi_p(f^{l})+b^l),
\end{eqnarray}
where $\varphi_p()$ is a many-to-one mapping with a defined operation corresponding to max- or average- pooling.

\begin{figure}[t]
  \centering
  \includegraphics[width=1.0\linewidth]{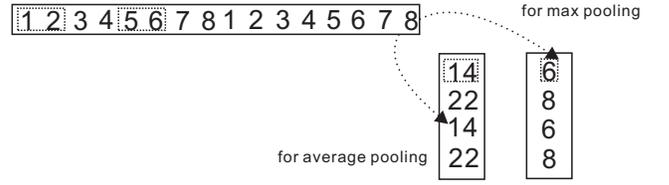}
  \caption{Strategy to vectorize pooling. Illustration of pooling for one 4x4 feature map.}
  \label{fig:fig4} 
\end{figure}

Due to the information loss, the inverse operator $\varphi^{-1}_p()$ is not well defined. We simply use nearest neighbor upscaling for approximation during backpropagation.

The pooling operations, both average pooling and max pooling, can be
thought of as a vector accumulation process guided by a pre-defined
index map, which can be similarly implemented by
\textit{accumarray()}. The only difference for max pooling is a
\textit{max} function is involved. For overlapping pooling, we could
insert the overlapped elements into the feature map and then apply
the same pooling strategy.
%
%

\subsection{Vectorizing Fully Connected Layers}

The fully connected layers (i.e. ``{\bf d}" in Fig. \ref{fig:fig1}), can be written in a dense matrix-matrix multiplication. Thus both the feed forward and backpropagation are naturally vectorized, in a unified matrix-matrix form.

\subsection{Vectorization for Mini-batches}
Our vectorization strategy can be directly extended to support
mini-batch training. Given a batch of samples indexed by $j$, the mini-batch training with a convolution layer is given by
\begin{eqnarray}
[f_i^{l+1}]_{i} = \sigma([\varphi_c(f^{l}_{,j})]_j[w_i^l]_i+[b_i^l]_i),
\end{eqnarray}
where $[~]_j$ is to assemble the matrix of different samples.

\begin{figure}[t]
  \centering
  \includegraphics[width=1.0\linewidth]{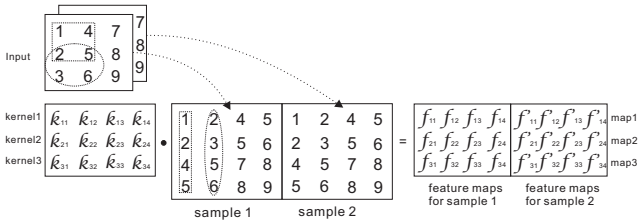}
  \caption{Strategy to vectorize mini-batch operations.}
  \label{fig:fig5} 
\end{figure}

Figure \ref{fig:fig5} shows the Matlab implementation of batch mode for the same operation as
in Fig. \ref{fig:fig2} with the batch size of 2. Both samples in the input batch are vectorized and the
outputs are arranged horizontally. We can show that the product of
the same kernel matrix as in Fig. \ref{fig:fig2} and this matrix is able to simultaneously generate feature maps for both
samples. Note that if an input sample of a convolutional layer has
multiple channels, we could treat it as a multi-channel feature map as shown in Fig. \ref{fig:fig3}.

%
\section{Experiments and Analysis}
The goal of the experiments presented in this section is to understand
the role of vectorization in training and testing CNNs as well as its limitation.

In order to make our experiment results of high validity and
relevancy, we compared the training and testing speed of our fully vectorized
implementation with Caffe and Cudaconvnet, the speed is competitive,
if not faster in all the tested cases.

\subsection{Comparing Different Degrees of Vectorization}
In this section, we seek to understand the role of vectorization by comparing six CNN implementations. These implementations differ in the degree of vectorization, which is illustrated in table 1. Imp-1 is a least vectorized one, Imp-2 naively parallelize the process of batch samples by adding a parallel for-loop to Imp-1 whilst Imp-6 is a fully vectorized implementation guided by the approaches we introduced. When working with one particular implementation we also observe how results change with network scales. The reason is we would like to examine different vectorization strategies with both small scale network and large scale ones and we are particularly interested in large scale CNNs since it is more relevant to recent advances in the field.

We consider three scales. Scale 1 is a small model, it is very similar to the standard LeNet \cite{lecun98}, but with more feature maps in the convolutional layers\footnote{This setting of conv layer is the same in Caffe's MNIST example. We have 2 fully connected hidden layers.}, ReLU nonlinearity and cross-entropy error. Scale 2 is a large network with about 60 million trainable parameters which is comparable to AlexNet \cite{KrizhevskySH12}. However, the architecture is tailored for the purpose of this study. First, we would like to keep the number of conv layer and the number of fully connected layer balanced so that we shall have a fair performance breakdown. It also allows us to directly compare the results with Scale 1. Second, to enable a fair comparison, we would like to have a unified vectorization scheme for convolution throughout the network. Thus unlike AlexNet which uses a stride 4 convolution in the first conv layer and stride 1 thereafter, all convolution operations use the same stride of 1 in our model. The consequences of those are, compare to AlexNet, scale 2 tends to have more feature maps in the conv layer but smaller size for the input images. We set the number of output units to 1000 as in AlexNet. Scale 3 is a larger network with 10,000 output units. This pushes the number of trainable parameters to 94 million, keeping other settings the same as scale 2. The performance on GPU\footnote{GeForce GTX 780 Ti, same card was used in the rest of the experiments.} (in terms of the number of images to be processed per second) of the six CNNs with different network scales during training is illustrated in table 2.

\begin{table}[t]
\centering \small
\begin{tabular}{l*{6}{c}r}
Vec ele          & Fu-co & Conv & Pool & Feat & Batch\\
\hline
Imp-6            & \checkmark & \checkmark & \checkmark & \checkmark & \checkmark  \\
Imp-5            & \checkmark & \checkmark & \checkmark & \checkmark   \\
Imp-4            & \checkmark & \checkmark & \checkmark  \\
Imp-3            & \checkmark & \checkmark   \\
Imp-2            & \checkmark &  &  &  & \checkmark  \\
Imp-1            & \checkmark
\end{tabular}
\caption{Different vectorized elements included in the six CNNs.
Fu-co: fully connected layer, Conv: convolutional layer, Pool:
pooling layer, Feat: feature map pachification, Batch: vectorize for
batch. A tick indicates the element is vectorized}
\label{tab:table1}
\end{table}

We are able to observe several insights from the figures. First of all, the results indicates that vectorization is vital to CNN's speed. We can see that Imp-1 is very slow and a naive parallelization (Imp-2) seems work poorly on GPU. Especially for the large scale networks, Imp-1 and Imp-2 are simply too slow to be practical. When training a small network, a fully vectorized CNN (Imp-6) is more than 200 times faster than the naive parallelization version during training and more than 100 times faster during testing. This acceleration is going to be more significant for bigger networks since Imp-1 and 2 scale poorly.

Second, all vectorization element we introduced contribute significantly to the final performance, during both training and testing. One interesting insight is the contribution of vectorizing pooling and feature map patchification seems to increase with the scale of the network. For instance, in table 2, Imp-4 (vectorize for pooling) has a 1.9x speed up than Imp-3 under scale 1 but a 4.5x speed up and a 4.3x speed up under scale 2 and scale 3 respectively. Same phenomenon happens for testing. This strongly indicates that the vectorization strategy for those two elements scales well with the size of the network.

\begin{table}[!htb]
\centering \small
\begin{tabular}{l*{6}{c}r}
\#img/sec          & Imp-1 & Imp-2 & Imp-3 & Imp-4 & Imp-5 & Imp-6 \\
\hline
Scale 1            & 1 & 6.1 & 15.3 & 29.5 & 85.4  & 1312.1 \\
Scale 2            & n/a & n/a & 2.4 & 11 & 42.3 & 188.7 \\
Scale 3            & n/a & n/a & 2.3 & 10 & 34.3 & 161.2 \\
\multicolumn{7}{l}{%
  \begin{minipage}{5.5cm}%
    \scriptsize $*$ batch size is 100 for scale 1 and 200 for scale 2 and 3.%
  \end{minipage}%
}\\
\end{tabular}
\caption{Training performance of the six CNNs (\#images to be
processed per second). Scale1: small model, 10 output units; Scale2:
large model, 1000 output units, Scale3: larger model, 10000 output
units.} \label{tab:table2}
\end{table}

On the other hand, we also observe that vectorizing batch processing brings more than 10x speed up for small models but only 3x to 5x speed up for large scale models. The contribution of vectorizing batch processing to the performance seems to decrease when scaling up the network though the speed up remains significant. We further investigate this phenomenon in the next section which leads to a strategy to achieve optimal training and testing speed.

\subsection{In Search of Optimal Speed}
We investigated the puzzle of decelerating speed up by scrutinizing the performance against different batch sizes. The results are presented in table 3 and 4 for training and testing respectively.


\begin{table}[!htb]
\centering
\small
\begin{tabular}{l*{6}{c}r}
\#img/sec          & b=1 & b=100 & b=200 & b=300 & b=400 \\
\hline
Scale 1            & 88.5 & 1312 & 1450.9 & 1574.2 & 1632.8 \\
Scale 2            & 41.9 & 136.9 & 188.7 & 192.3 & 106.3 \\
Scale 3            & 34.3 & 123.5 & 161.3 & 163.9 & 91 \\

\end{tabular}
\caption{Training performance of Imp-6 against different batch sizes (\#images to be processed per second).}
\label{tab:table4}
\end{table}

\begin{table}[!htb]
\centering
\small
\begin{tabular}{l*{6}{c}r}
\#img/sec          & b=1 & b=100 & b=200 & b=400 & b=600 \\
\hline
Scale 1            & 151.5 & 1812.6 & 1878.4 & 2023.5 & 2192.2 \\
Scale 2            & 75.8 & 222.2 & 270.2 & 285.7 & 103.1 \\
Scale 3            & 74 & 212.8 & 256.4 & 277.8 & 89.2 \\

\end{tabular}
\caption{Test performance of Imp-6 against different batch sizes (\#images to be processed per second).}
\label{tab:table5}
\end{table}

In table 3, we can see that for the small model (scale 1) the acceleration brought by each adjacent batch size increase is 14x, 1.1x, 1.08x and 1.03x. The acceleration obtained via the increase of batch size seems to be rapidly vanishing. For the large model (scale 2), the first three acceleration ratio are 3.2x, 1.3x and 1.02x, demonstrating the same vanishing trend. Further increase in batch size even leads to a performance degradation instead. Same situation occurs for the larger model (scale 3). Though the ability of processing 192 images/second for training and 285 images/second for testing with our commodity GPU for the scale 2 network is promising, this result still indicates that there is some scaling limitation within the vectorization for batch processing. Similar results in table 4 seems to further suggest that such limitation is shared between training and testing. In order to completely understand the rationale under the hood, we have to resort to a detailed performance breakdown.

\subsubsection{Performance Breakdown and Limitation. }
We decompose the whole training procedure into the following components. They are 1) conv layers; 2) pooling layers; 3) fully connected layers; 4) others (e.g. ReLU, cost). We distinguish the statistics between forward pass and back-propagation, therefore 8 components to look at.


\begin{figure}[t]
  \centering
\begin{tabular}{@{\hspace{0mm}}c@{\hspace{1mm}}c@{\hspace{1mm}}c@{\hspace{1mm}}c@{\hspace{1mm}}c}
\includegraphics[width=0.35\linewidth]{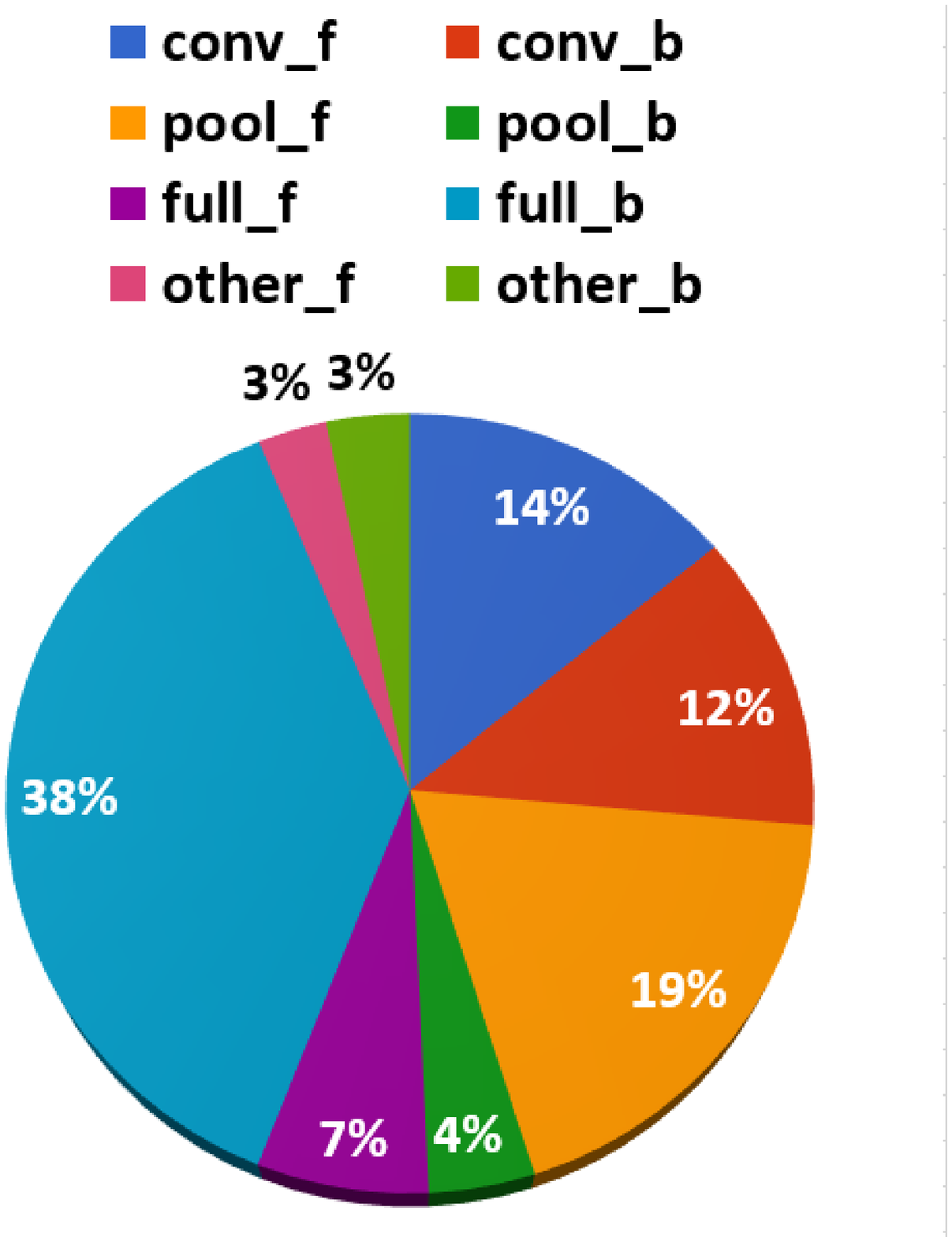}&
\includegraphics[width=0.35\linewidth]{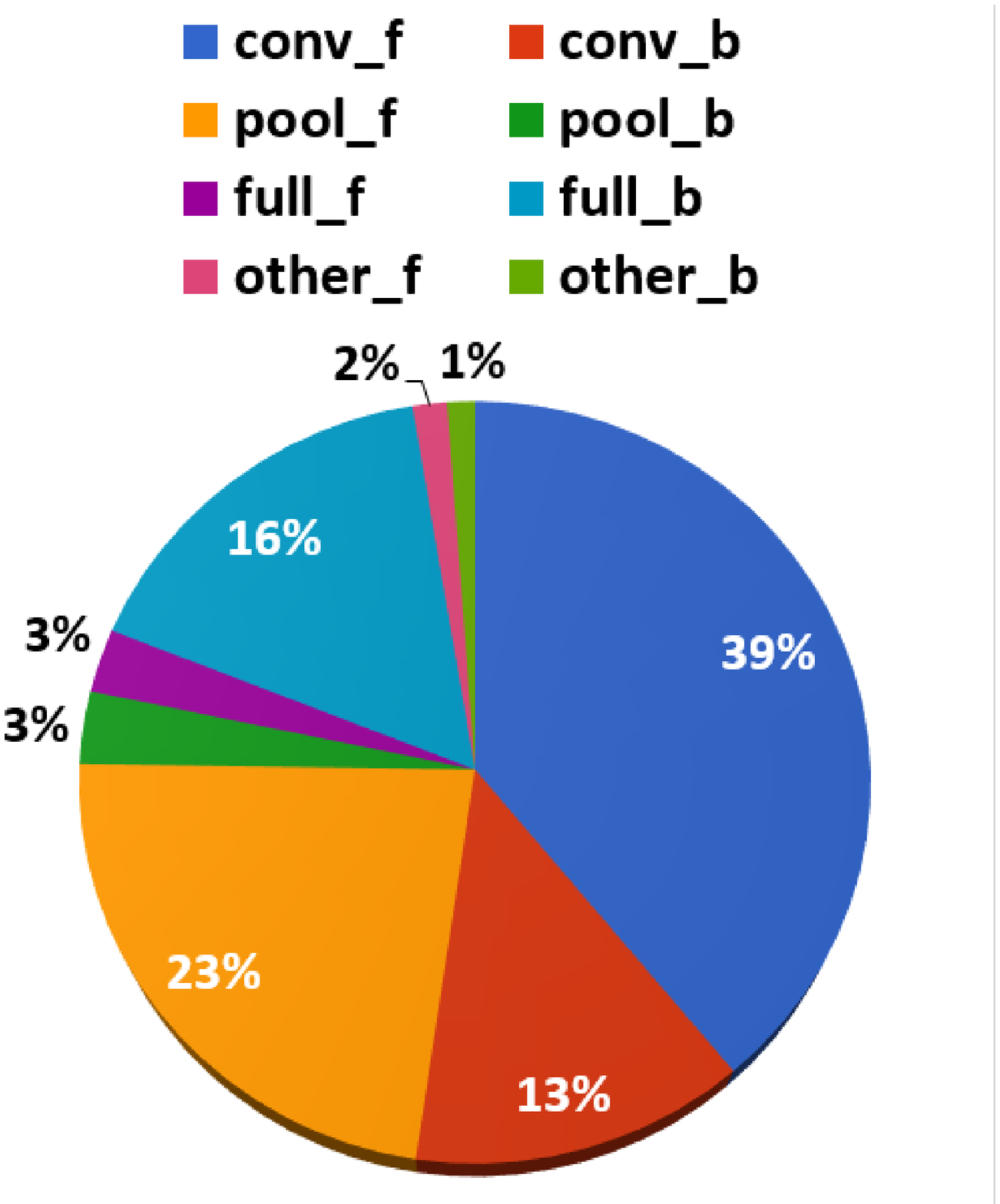}&
\includegraphics[width=0.35\linewidth]{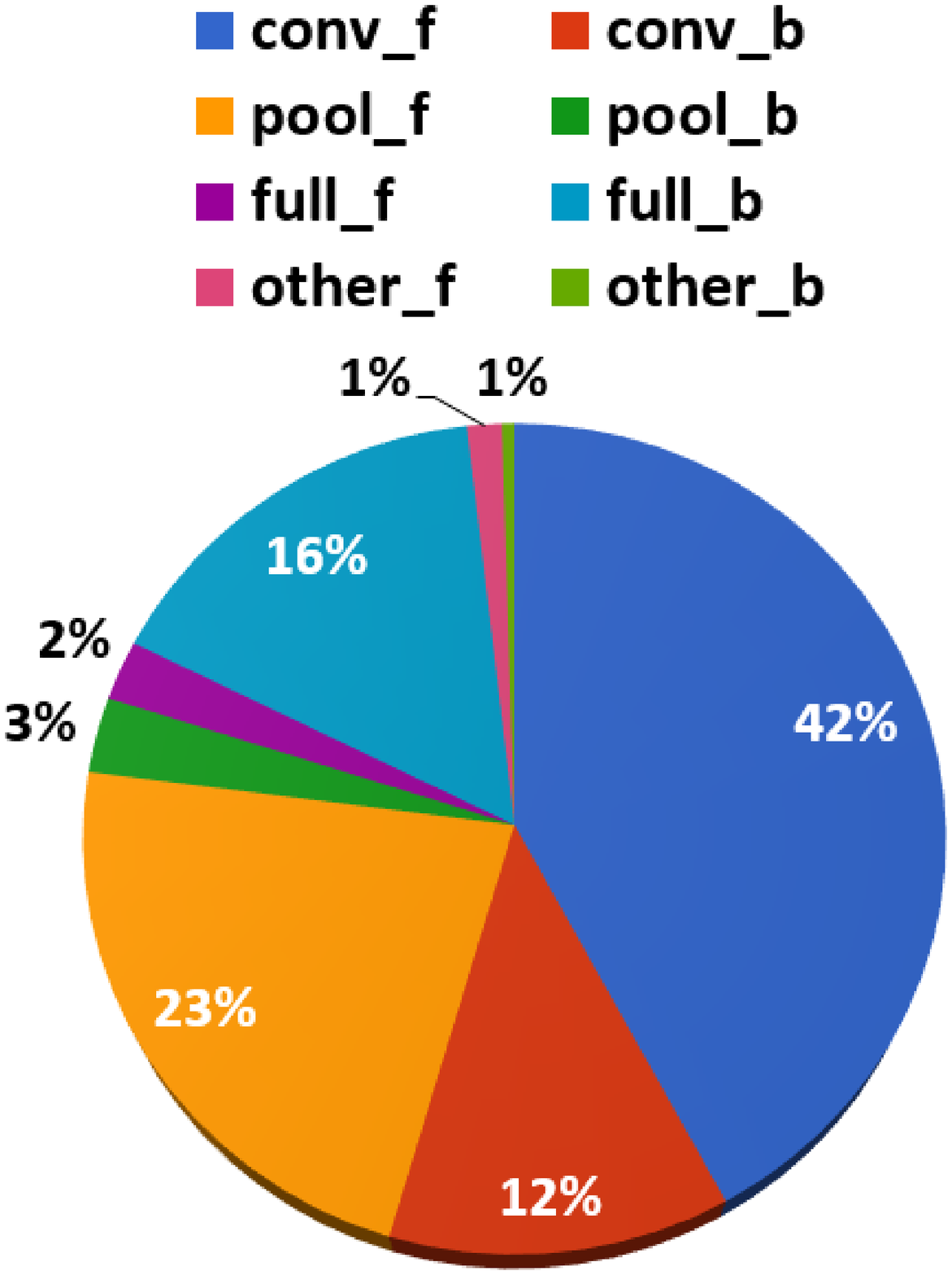}
\\
(a) & (b) & (c)
\end{tabular}
  \caption{Performance break down. (a) Scale 3 network, batch size = 1. (b) Scale 3 network, batch size = 200. (c) Scale 3 network, batch size = 300. \textit{conv}: conv layers, \textit{pool}: pooling layers, \textit{full}: fully connected layers, \textit{other}: other operations, \textit{\_f}: forward pass, \textit{\_b}: back-propagation. }
  \label{fig:fig6} 
\end{figure}

Figure 6 illustrates the performance break down (in terms of the
proportion of computing time in processing one batch) during
training of the two representative cases from our largest network
(scale 3) in the experiment. Batch size is 1 for Fig. 6(a), 200 for
Fig. 6(b) and 300 for Fig. 6(c). We can observe from Fig. 6(a) that
44\% of the overall time was used in processing the fully connected
layers. It was in fact the biggest consumer of the computing time
for this batch size. We also see that the time spent on full\_b is
significantly more than full\_f. This makes sense because it
involves larger matrix multiplication and larger transform matrix
than that in the forward pass. The second largest consumer of time
is the convolution layers and we can see that the time spent in
forward pass and back-propagation is reasonably balanced.

However, the situation we found in Fig. 6(b) and Fig. 6(c) is very
different. One obvious character is, when increasing the batch size,
the time costs by the conv layers is now considerably more than the
fully connected layers. While the proportion between full\_f and
full\_b among the three batch sizes roughly remains the same, we
found conv\_f spent much more time than conv\_b for large batch
sizes. This indicates the scaling limitation is within the conv\_f
when vectorizing for batch processing. A further scrutiny on this
issue shows that the limitation is caused by the following two
factors namely the memory overhead in handling multiple samples and
the overhead caused by invoking patchification on bigger samples.
While there might be alternative strategies to vectorize batch
processing, we argue that the aforementioned overhead is hard to be
completely avoided.

\subsubsection{Finding the Optimal Speed. }
We found the observations from Fig. 6 are also valid for scale 1 and
scale 2 networks, but with an important difference. For small
networks like the scale 1 network, the acceleration brought by batch
processing shall be valid for very big batch sizes (e.g. 1000)
whilst for large networks batch size needs to be chosen carefully or
else the speed degradation like we saw in table 3 and 4 shall occur
before the network hits the GPU memory ceiling. This suggests that
given a network design choosing an appropriate batch size may be
vital in achieving the optimal speed. Based on our scale 2 network,
we select 10 other networks by randomly adjusting several parameters
such as filter size, number of feature maps, number of output units
and sigmoid function, etc. We run these networks for both training
and testing by adjusting the batch sizes to see if this contention
is generally applicable for large networks.


\begin{figure}[t]
  \centering
\begin{tabular}{@{\hspace{0mm}}c@{\hspace{1mm}}c@{\hspace{1mm}}c@{\hspace{1mm}}c@{\hspace{1mm}}c}
\includegraphics[width=0.5\linewidth]{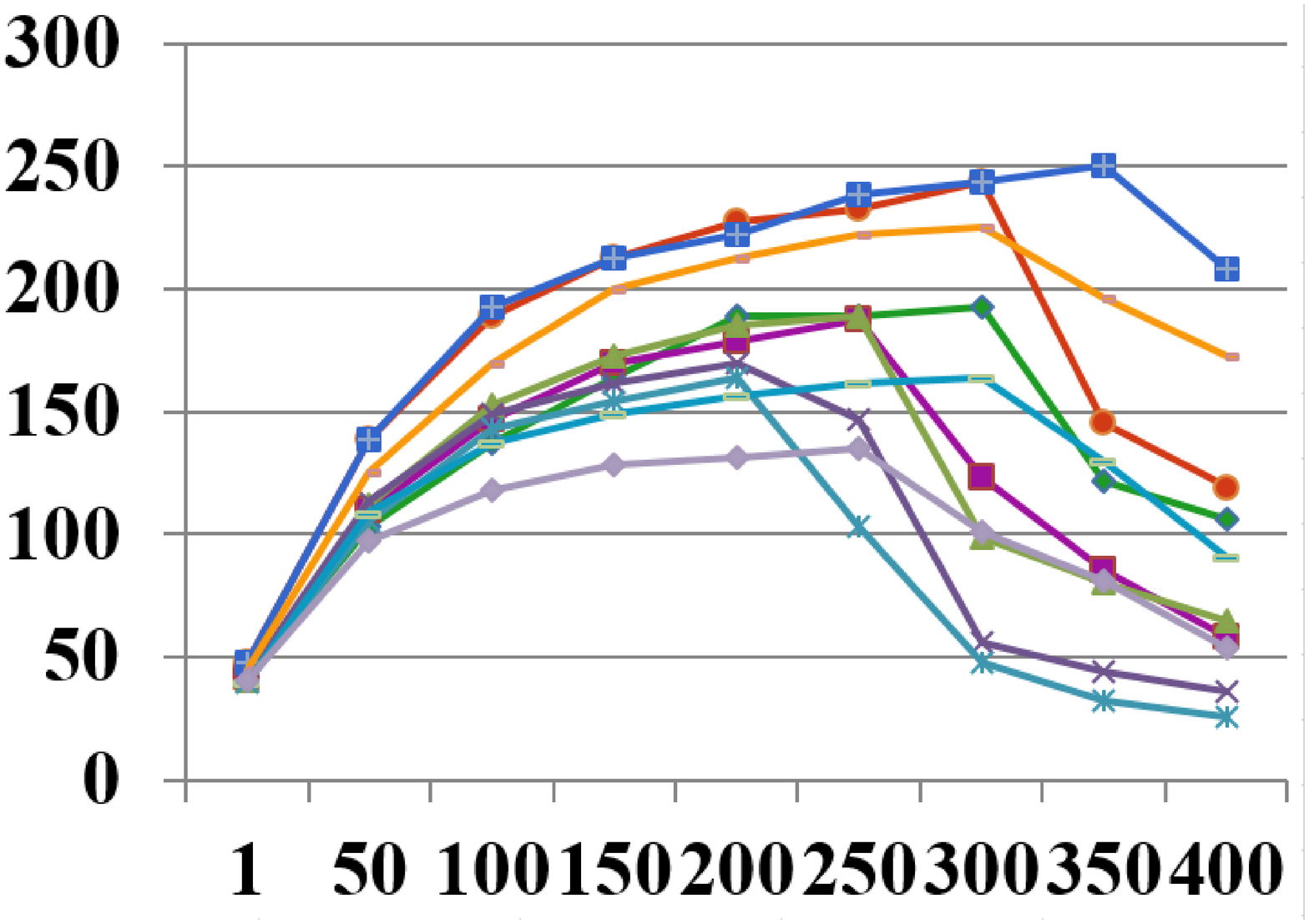}&
\includegraphics[width=0.5\linewidth]{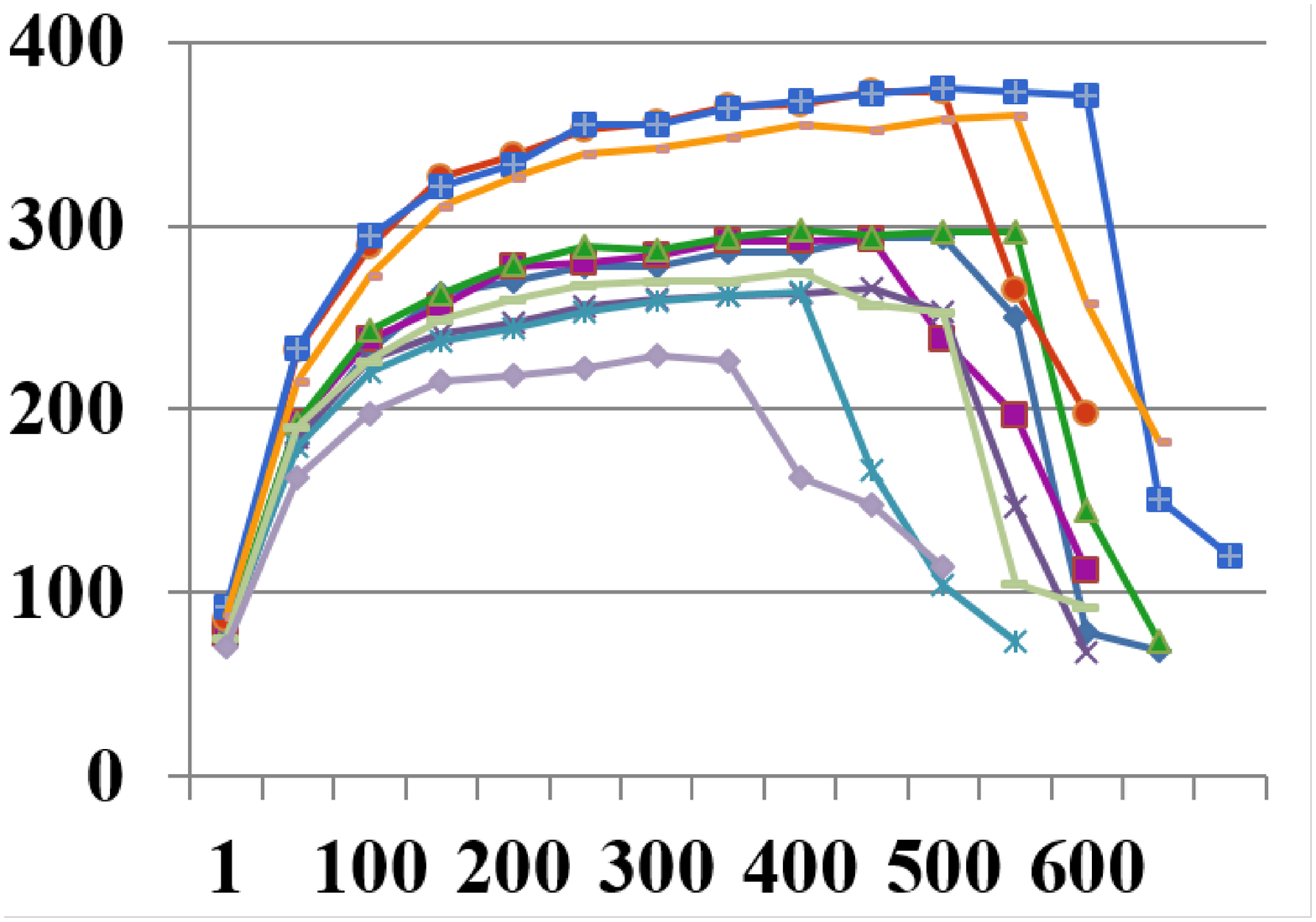}\\
(a) & (b)
\end{tabular}
  \caption{Speed of 10 randomly selected networks. X axis, batch size. Y axis, number of images to be processed per second. (a) for training. (b) for testing.}
  \label{fig:fig7} 
\end{figure}
Figure 7 confirms our aforementioned contention for large
networks and makes the importance of choosing an appropriate batch size
obvious. First, it suggests that the optimal batch size among
different network parameters is usually quite different. Directly
adopting a batch size from a previous set of network parameters may
lead to significantly inferior speed. Second, it also suggests
that the optimal batch size between the training stage and the
testing stage is also different, even if for the same network. A
naive adoption of the batch size from the training stage is often
not optimal and leads to considerable speed loss. These findings has
direct implications in building real-time systems in which
optimization for model testing is the key.

\section{Unification of High{/}Low Level Vision Tasks }
Despite the rapid adoption of deep CNN in addressing various kinds of high level computer vision tasks typified by image classification and object localization, other problems such as detecting objects of different shapes in real-time seem still a problem under investigation. On the other hand, we observed that there are a few very recent studies \cite{xu14,Eigen14} successfully used deep CNN in various low level vision tasks such as image deblurring and denoising, etc. Though the domain knowledge required to build those new networks substantially differ from that used in addressing high level vision tasks, same vectorization principles presented in this paper will apply.

More interestingly, the same vectorization principle across those tasks actually gives us a chance (perhaps for the first time) to unify both high level vision tasks and low level vision tasks in a single computational framework. In this section, we introduce the application of our VCNN implementation in tasks seemingly of distinct fields namely, image denoising and deblurring (low level vision) as well as multi-object detection (high level vision).

\subsection{CNN for Image Processing }
\begin{figure}[!htb]
  \centering
    \includegraphics[width=1.0\linewidth]{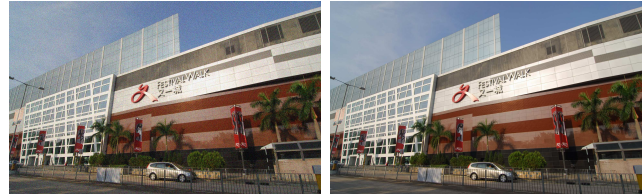}
  \caption{Application in image denoising. }
  \label{fig:fig8} 
\end{figure}

\begin{figure}[!htb]
  \centering
    \includegraphics[width=1.0\linewidth]{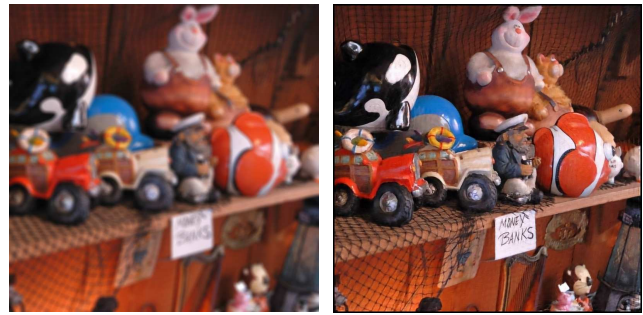}
  \caption{Application in image deconvolution. }
  \label{fig:fig9} 
\end{figure}

Image processing tasks do not require pooling and fully connected layers in general.  To verify the effectiveness of the proposed vectorized framework, we implemented a network architecture by simply removing the pooling and fully connected layers from Fig. \ref{fig:fig1} and trained the network with synthesized clear-noisy image pairs. One of the denoise result is given in Fig. \ref{fig:fig8}. Another sample application of our vectorized CNN is the recent proposed image deconvolution \cite{xu14}. Result is shown in Fig. \ref{fig:fig9}.


\subsection{Novel Training Scheme for Multi-object Detection }
Conventional image classifiers are usually trained by image samples with equal sizes. This imposes a critical limitation when applying it in detection. For instance, it is reasonable to put a human face sample in a square image, but doing so for non-squared objects (e.g. shoes) tends to include more background content thus introduces more noise which is detrimental to accurate and efficient object detection. One possible alternative is to formulate object detection as a regression problem \cite{szegedy13}, however, it requires a very large amount of data and usually very big models to capture the variety of the possible patterns.

\begin{figure}[!htb]
  \centering
    \includegraphics[width=1.0\linewidth]{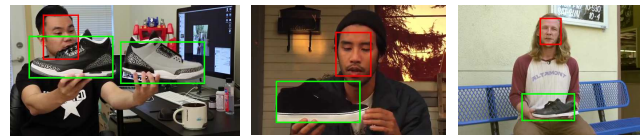}
  \caption{Application in real time multi-object detection. Shoes review videos are from Youtube. }
  \label{fig:fig10} 
\end{figure}

Using VCNN, we were able to train a single image classifier but with heterogeneous input sizes by using vectorization. The key insight is heterogeneous inputs can actually share all the weights in a CNN except the ones in the connection between conv layer and fully connected layer. This approach not only avoids the background noise but also being a lot more lightweight than the regression approach. We successfully applied it in a detection system which runs in real-time. We can show that this approach tends to have less false alarms and works efficiently with multi-scale detection through vectorization.

\section{Conclusion}
In this paper, we elaborate several aspects on vectorization of
deep CNN. First, we present the vectorization steps of all essential parts of implementing deep CNNs. The vectorization steps are further exemplified by Matlab practices. Second, we have developed and compared six CNN implementations
with different degrees of vectorization to analysis the impact
of vectorization on speed. Third, based on the practices, we provide a unified framework for handling both low-level and high-level vision tasks. Experiments on various applications including image denoise, decovolution and real-time object detection demonstrated the effectiveness of the proposed strategies. As
the introduced vectorization techniques are general enough, our future
direction includes optimization for different hardware or cloud platforms.

\bibliographystyle{aaai}
\bibliography{vcnn}

\end{document}